\documentclass{article}

\usepackage{microtype}
\usepackage{graphicx}
\usepackage{subcaption}
\usepackage{booktabs} % for professional tables

\usepackage{hyperref}
\usepackage{tikz}
\usetikzlibrary{
  arrows.meta,
  positioning,
  fit,
  calc,
  shapes.geometric,
  decorations.pathreplacing
}
\usepackage{xcolor}

 \usepackage[accepted]{icml2026} 

\usepackage{amsmath}
\usepackage{amssymb}
\usepackage{mathtools}
\usepackage{amsthm}

\usepackage[capitalize,noabbrev]{cleveref}

\theoremstyle{plain}

\theoremstyle{definition}

\theoremstyle{remark}

\usepackage[textsize=tiny]{todonotes}

\icmltitlerunning{Position: Weight Space Should Be a First-Class Generative AI Modality}

\begin{document}

\twocolumn[
  \icmltitle{Position: Weight Space Should Be a First-Class Generative AI Modality}



  \icmlsetsymbol{equal}{*}

  \begin{icmlauthorlist}
    \icmlauthor{Zhangyang Wang}{yyy}
    \icmlauthor{Peihao Wang}{yyy}
    \icmlauthor{Kai Wang}{comp}
  \end{icmlauthorlist}

  \icmlaffiliation{yyy}{University of Texas at Austin}
  \icmlaffiliation{comp}{Tencent Hy}

  \icmlcorrespondingauthor{Zhangyang ``Atlas" Wang}{atlaswang@utexas.edu}

  \icmlkeywords{Machine Learning, ICML}

  \vskip 0.3in
]



\printAffiliationsAndNotice{}  % no special notice (required even if empty)

\begin{abstract}
Neural network checkpoints have quietly become a large-scale data
resource: millions of trained weight vectors now exist, each encoding
task-, domain-, and architecture-specific knowledge.
This position paper argues that \textit{model checkpoints should be treated as a
first-class data modality, and that generative modeling in weight space
should be standardized as a core machine learning primitive}.
Recent advances demonstrate that neural weights
can be synthesized on demand, often matching fine-tuning performance
while reducing adaptation cost by orders of magnitude.
We contend that these results reflect an underlying structural fact:
high-performing models occupy low-dimensional, highly structured regions
of weight space shaped by symmetry, flatness, modularity, and shared
subspaces.
Building on this view, we organize existing methods into a five-stage pipeline, survey
applications where the approach is already practical, and clarify current
limits: adapter-scale and conditional generation are advancing rapidly,
while unrestricted frontier-scale checkpoint synthesis remains open. Our goal is to shift the community’s default mindset from optimizing
models per task to sampling models from learned weight distributions, accelerating toward an era in which \textit{AI systems
routinely improve or create other AI systems}.

\vspace{-1em}
\end{abstract}

\section{Introduction and Statement of Position}
\label{sec:intro}

\textit{The success of generative AI has so far been asymmetric: we can
\emph{generate} pixels, waveforms, and tokens on demand, but we still
\emph{construct} - via repeated, costly optimization - the models that
perform this generation.}
This asymmetry increasingly defines the bottleneck of modern machine
learning.
Training a single foundation model can emit as much
$\mathrm{CO}_2$ as a small airline fleet~\cite{patterson2021carbon},
while even modest fine-tuning remains prohibitive for many start-ups,
academic groups, and edge deployments.
At the same time, demand for
\emph{ultra-rapid}, \emph{task-specialized}, and
\emph{user-personalized} models is accelerating—from on-device copilots
to adaptive robotics to personalized healthcare.

We argue that progress is throttled less by hardware than by
\emph{methodological inertia}.
Neural network parameters are still treated as immutable artifacts to
be rediscovered from scratch for every new task or user.
Yet a century of statistics 
teaches a different lesson: when data exhibit structure, we should
\emph{model and sample} that structure rather than repeatedly re-optimizing it.
Recent results indicate that trained neural weights themselves form such
a structured data modality.
HyperNetworks generate full convolutional weight tensors in a single
forward pass~\cite{ha2017hypernetworks};
graph-conditioned predictors initialize unseen architectures without
task-specific training~\cite{knyazev2021parameter};
diffusion models denoise checkpoints in weight space to yield
ImageNet-ready ConvNeXt backbones~\cite{wang2024neural,wang2024rpg}.
More recently, \cite{liang2025drag} demonstrates that even large language models (LLMs) can be adapted by
\emph{directly generating} low-rank weight updates in seconds,
often matching or exceeding fine-tuning performance.

\begin{figure*}[htp]
    \centering
    \includegraphics[width=.95\linewidth]{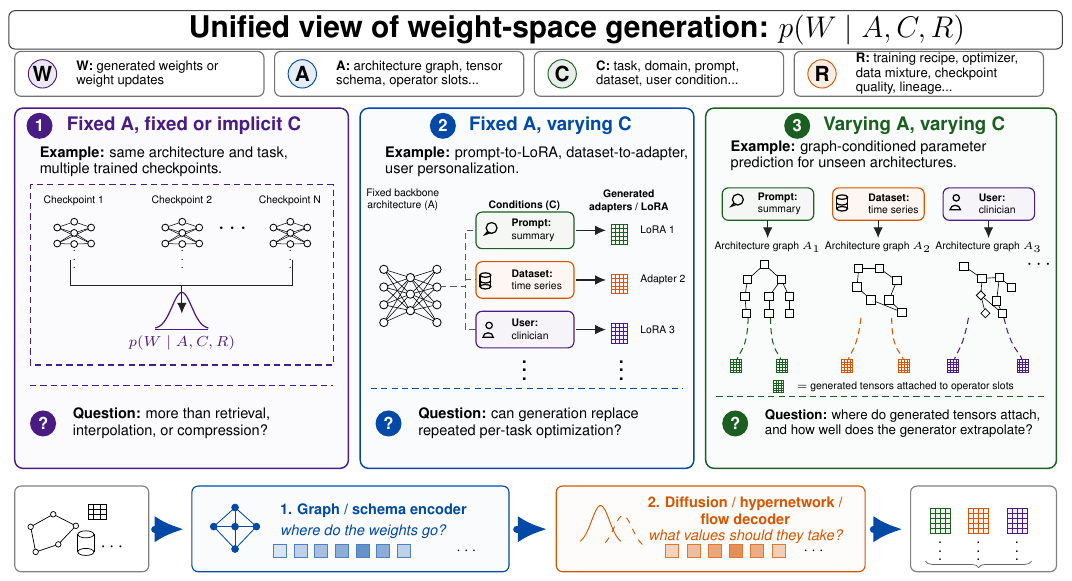}
    \caption{
   A regime map for weight-space generation (see \$3). We frame neural weight generation as conditional sampling from \(p(W \mid A,C,R)\), where \(W\) denotes generated weights or weight updates, \(A\) specifies the architecture graph and tensor schema, \(C\) encodes task or user conditions, and \(R\) captures training-recipe and checkpoint-lineage information. The three regimes distinguish whether architecture and conditioning are fixed or varied: checkpoint-level generation under fixed settings, condition-driven generation for adapters or personalization, and graph-conditioned generation across architectures. This view separates the placement problem, handled naturally by graph or schema encoders, from the value-generation problem, handled by diffusion, hypernetwork, or flow decoders.}
    \label{fig:achievements_and_promise}    \vspace{-0.9em}
\end{figure*}

\vspace{-0.7em}\paragraph{Position.}
We posit that \textit{\textbf{neural network checkpoints constitute a
first-class data modality, and that generative modeling in weight space
should be standardized as a core machine learning primitive.}}
Under this view, model creation is reframed as
\emph{conditional sampling}: given a task description, data domain,
user context, or architectural specification, one samples from a learned
distribution of high-performing weights. Crucially, this position does \emph{not} claim that weight generation
solves all problems of training or scalability.
Nor is it reducible to the connectedness of minima
\cite{entezari2021role,garipov2018loss}:
connectivity guarantees that paths exist, but it does not explain how to
reliably reach flat, robust, calibrated, or domain-adapted regions.
Achieving those properties requires learning an explicit,
conditioned density over weight configurations.

\vspace{-0.7em}\paragraph{Why Now?}
Weight-space generation is no longer a speculative idea; it is enabled
by a concrete shift in the ML ecosystem.
Open-science efforts have produced an unprecedented corpus of trained
models:
(i) the Hugging Face Hub alone hosts over one million public checkpoints
across modalities;
(ii) TensorFlow Hub, ONNX Model Zoo, and open releases from major
industrial labs further expand this repository;
(iii) recent academic studies explicitly treat neural weights as data,
including large-scale analyses of tens of thousands of trained models;
(iv) parameter-efficient fine-tuning has exploded the number of
\emph{task-conditioned} checkpoints, with over 22\,000 LoRA adapters for
the LLaMA family alone as of 2025.
Together, these trends provide both the \emph{data substrate} and the
\emph{deployment pressure} that make generative modeling in weight space
timely and actionable.

\vspace{-0.5em}\paragraph{Contributions.}
This paper advances the position through four concrete contributions:
\begin{enumerate}\vspace{-0.5em}
    \item \textbf{Theory.}
    We synthesize decades of optimization theory and empirical results, to explain why generative modeling of weights is
    both feasible and non-trivial.\vspace{-0.5em}
    \item \textbf{Framework.}
     We formulate weight-space generation as conditional sampling
    \(p(W \mid A,C,R)\) (Fig \ref{fig:achievements_and_promise}), distinguishing fixed-architecture generation,
    condition-driven adaptation, and cross-architecture generation, and
    relate these regimes to a five-stage pipeline:
    tokenization \(\rightarrow\) embedding \(\rightarrow\) generative
    predictor \(\rightarrow\) training strategy \(\rightarrow\) evaluation.\vspace{-0.5em}
    \item \textbf{Evidence.}
    We survey demonstrated applications, while separating
    practical current regimes from still-open frontier-scale full
    checkpoint generation.\vspace{-0.5em}
    \item \textbf{Agenda.}
    We analyze alternative views, failure modes, and
 concrete steps toward this new paradigm.\vspace{-1em}
\end{enumerate}
By elevating weights from optimization artifacts to generative objects,
we aim to shift the community’s default question from
\emph{“How do we fine-tune this model?”} to
\emph{“What distribution of models should we sample from?”}

\section{Theoretical Foundation of Weight Space Geometry:
Why Feasible, Yet Non-Trivial}
\label{sec:geometry}

If neural network weights are to be treated as a \emph{generative
modality}, then any successful generator must respect the intrinsic
geometry of weight space itself.
This section synthesizes what decades of optimization theory and recent
empirical studies reveal about that geometry, and explains how they both \emph{enable} and \emph{constrain} weight 
generation.

\textit{Far from forming a random high-dimensional cloud}, the set of
high-performing neural networks concentrates on structured subsets of
parameter space.
These structures explain why learned generative models can empirically succeed.

\vspace{-0.5em}\subsection{Mode Connectivity of Solutions}\vspace{-0.2em}

Partly because of over-parameterization, many trained neural networks do not behave as isolated optima in the raw parameter space. While naive linear interpolation between independently trained models
often incurs large loss barriers~\cite{shevchenko2020landscape,kuditipudi2019explaining},
empirical results revealed \emph{mode connectivity}:
continuous low-loss paths linking solutions trained from different
initializations~\cite{garipov2018loss,draxler2018essentially}.

Theoretical work clarifies when such connectivity arises.
Solutions exhibiting dropout or noise stability admit barrier-free
connections in ReLU networks~\cite{kuditipudi2019explaining}, and
sufficiently wide networks provably admit near-zero-loss paths between
SGD solutions~\cite{shevchenko2020landscape}.
Even moderately over-parameterized architectures, such as CIFAR-10
ResNets, empirically exhibit this phenomenon~\cite{garipov2018loss,draxler2018essentially}.

Connectivity provides evidence that many low-training-loss SGD solutions are not isolated in parameter space. It does not by itself prove that all well-generalizing solutions form a single manifold. 
Furthermore, this result alone does \emph{not} make weight generation
trivial.
Connectivity guarantees that \emph{paths} exist, but offers no mechanism
for \emph{sampling} points with desired properties such as robustness,
calibration, or transferability.
For generation, the challenge is hence not reachability, but
\emph{density modeling} (see \$2.4): reachability between trained points is often possible, but sampling robust, calibrated, transferable, or safe models still requires learning a structured conditional density.

\vspace{-0.5em}\subsection{Permutation Symmetries and Quotient Geometry}\vspace{-0.2em}

A central reason for the apparent multiplicity of minima is parameter
symmetry~\cite{goodfellow2014qualitatively,choromanska2015loss}.
Permuting hidden units or channels preserves the network’s function
while producing numerically distinct weight vectors.
As a result, each functional solution corresponds to an equivalence
class in raw parameter space, creating flat directions and degenerate
minima.

Interpolation between unaligned solutions introduces artificial loss
barriers due to mismatched neuron identities.
When permutation symmetry is explicitly factored out, these barriers
vanish.
\cite{entezari2021role} conjectured that, modulo permutation, SGD
solutions occupy a single basin.
\textit{Git Re-Basin}~\cite{ainsworth2022git} provided constructive
evidence, demonstrating zero-barrier linear connectivity after neuron
alignment.
Related symmetry-aware matching improves model averaging and federated
learning~\cite{yurochkin2019bayesian,tatro2020optimizing,zhao2024texttt}.

For generative modeling, this implies that the natural object of study
is not raw weight space $\mathbb{R}^n$, but a \emph{quotient space} under
symmetry.
Generators that ignore this structure waste capacity modeling redundant
degrees of freedom.

\vspace{-0.5em}\subsection{Flatness and Low Intrinsic Dimension}\vspace{-0.2em}

Beyond symmetry, trained solutions occupy remarkably flat regions of
the loss landscape.
Hessian analyses show that only a small number of eigenvalues are large
at SGD minima, with most directions exhibiting near-zero
curvature~\cite{sagun2018,ghorbani2019investigation}.
Perturbations along flat directions often leave performance
unchanged. Flatness correlates strongly with generalization~\cite{jiang2019fantastic},
and invariant measures confirm that broader basins generalize better
despite reparameterization issues~\cite{dinh2017sharp,petzka2021relative}.
Methods that explicitly promote flatness, such as Sharpness-Aware
Minimization~\cite{foret2021sam}, consistently improve performance.

Multiple lines of evidence indicate that the \emph{effective} dimension of optimization trajectories and searchable subspaces can be far smaller than the ambient parameter count.
Lottery Ticket and subspace training results show that networks can be
trained successfully within low-dimensional affine
subspaces~\cite{frankle2018lottery,chen2020lottery,li2018measuring,jaiswal2025}.
Large-scale training dynamics reveal that SGD trajectories
explore a tiny, low-rank subspace of parameter space
regardless of initialization~\cite{gur2018gradient,mao2024training}, though not without caveat whether a single dominant subspace really exists \cite{song2025does}.
Recent memory-efficient optimizer designs ~\cite{zhao2024galore,zhu2024apollo} explicitly exploit this phenomenon: see \cite{balzano2025overview} for more reviews. These results imply that high-performing weights occupy
a \emph{thin, structured manifold}.
Sampling must therefore capture strong inter-parameter correlations.

\vspace{-0.5em}\subsection{Implicit Bias: Why Solution Density Is Not Uniform}\vspace{-0.2em}

Optimizers exhibit strong implicit bias toward solutions with
specific geometric properties. This is critical for weight generation:
the target distribution is \emph{highly non-uniform}.
Effective generators must learn to model this density.

In deep linear and homogeneous networks, gradient descent with weight
decay provably minimizes effective rank over training
trajectories~\cite{ji2020implicit,le2022training}.
More broadly, SGD combined with common regularizers implicitly performs
rank minimization~\cite{galanti2024sgd}, explaining the low effective
rank observed. When data lie on a low-dimensional manifold, optimal solutions align
with that intrinsic structure~\cite{ongie2019function}.

As another example, SGD also favors small-norm, max-margin solutions in classification
settings~\cite{soudry2018implicit,lyu2019gradient}.
Although theory for nonlinear networks remains incomplete, empirical
evidence consistently shows that optimization concentrates probability
mass in a small subset of geometrically regular solutions.

\vspace{-0.5em}\subsection{Compositionality and Modularity}\vspace{-0.2em}
\label{sec:modularity}

Neural networks frequently internalize compositional structure when
trained on structured data.
Empirical studies show that modern networks decompose into coherent
sub-circuits corresponding to reusable functions or
concepts~\cite{csordas2020neural,qiu2023unlocking,zhang2023emergent},
with individual blocks exhibiting interpretable roles
\cite{bau2020understanding}. Recent theoretical and empirical work suggests that gradient-based
training induces latent modular representations within weight
space~\cite{cranmer2020discovering,zheng2022symbolic,chen2023symbolic,wang2025neural}.
Network morphisms such as Net2Net~\cite{chen2016net2net}
and progressive stacking~\cite{gong2019efficient,wang2023learning,wang2023data,yang2022deep}
explicitly exploit this modularity.
Model editing and merging~\cite{wortsman2022model,zhao2024texttt} further
demonstrate empirically that specific skills correspond to localized parameter
subsets.

For generation, modularity explains both promise and difficulty.
On one hand, generators can recombine learned modules to generalize
across tasks.
On the other, module boundaries are not smooth or spatially local in
raw parameter coordinates, complicating naive notions of locality and
requiring structured tokenization and curriculum strategies.

\vspace{-0.5em}\subsection{Shared Representations and Universal Subspaces?}\vspace{-0.2em}

Despite architectural diversity and/or training data or task differences, trained networks often converge toward
remarkably similar internal representations. The \emph{Platonic representation hypothesis}~\cite{huh2024platonic} was the first to
posit that high-performing models recover a common structure dictated
by the data-generating process rather than architectural details. More evidence includes linear alignment of representations across models
\cite{kornblith2019similarity} snd successful model stitching
\cite{bansal2021revisiting}.
Recent large-scale empirical studies suggest the existence of
shared low-dimensional subspaces that capture variation across datasets, initializations, tasks
and architectures \cite{kaushik2025universal}. Although still preliminary, this discovery has the potential to explain the success of transfer learning,
distillation, and adapter-based methods.
It also suggests that generators potentially need not learn
architecture-specific distributions from scratch, but can operate in
shared latent coordinates that instantiate across architectures.

\vspace{-1em}\paragraph{Summary.}
Taken together, 
neural weight space is structured and low-dimensional, biased by
optimization, making weight generation \emph{feasible}.
At the same time, the absence of smooth locality, the presence of
symmetry, and the non-uniform density make the problem
fundamentally non-trivial.
Any successful generator must therefore be symmetry-aware,
bias-aligned, and capable of capturing long-range correlations rather
than naive smoothness.

\section{Practical Mechanisms for Neural Weight Generation: A Standardized Taxonomy}
\label{sec:mechanisms-weight-gen}

Armed with the geometric insights of Section~\ref{sec:geometry}, we now
turn to the practical question: \emph{how can we learn a model that
samples from the structured and low-dimensional
distribution of high-performing neural weights?}
Contemporary systems increasingly converge on a modular pipeline that
mirrors the workflow of modern generative AI for images and text, but
with crucial differences: weights exhibit strong permutation symmetries, weak ``spatial locality'' in raw
coordinates, and heavy-tailed inter-layer dependencies.
As a result, each stage of the pipeline must be re-designed to encode
the right invariances and to avoid degenerate solutions.

In this section, we synthesize the rapidly growing literature into a standardized, 
five-stage blueprint: \textbf{tokenization},
\textbf{embedding}, \textbf{generative predictor}, \textbf{training
strategy}, and \textbf{evaluation}.
Our goal is \emph{mechanistic}: to explain how each
design choice operationalizes the geometry of Section~\ref{sec:geometry},
and to highlight which choices appear most practical today.

\vspace{-0.9em}\paragraph{Regimes and scope.}
To avoid overstating feasibility, we separate three regimes that recur
throughout this section:
(i) \emph{adapter generation} (e.g., LoRA-like low-rank updates),
(ii) \emph{mid-scale full-weight generation} (e.g., 10M--200M parameter
vision backbones), and
(iii) \emph{cross-architecture generalization} (predicting weights for
unseen graphs).
The field is currently most mature in (i) and increasingly competitive
in (ii), while (iii) remains an open frontier.

As depicted in Fig. \ref{fig:achievements_and_promise}, a useful unifying abstraction is the conditional family $p(W \mid A, C, R)$, 
where \(W\) denotes weights or weight updates, \(A\) denotes the architecture graph and tensor schema, \(C\) denotes task, domain, prompt, dataset, or user condition, and \(R\) denotes training recipe and quality metadata\footnote{In practice, \(R\) need not be a user-facing input: a user may specify only \(C\), while the generator marginalizes over, selects, or internally conditions on suitable recipe and quality factors.}. When \(A\) is fixed, the parameter slots are known and the generator mainly models values. When \(C\) varies, the generator performs conditional adaptation over a known layout. When \(A\) varies, the generator must also map generated tensors back to operator types, shapes, and connectivity. In this sense, graph encoders mainly solve the placement problem, while diffusion, flow, or hypernetwork modules solve the value-generation problem.

%---------------------------------------------------------------------
\vspace{-0.3em}
\subsection{Tokenization of Weights: What Is ``Local''?}
\label{sec:tokenisation}

Tokenization maps heterogeneous parameter tensors into a common
``language'' suitable for sequence, set, or graph models. Yet one must confront a key concern:
\emph{weight space lacks obvious locality}.
Unlike pixels or tokens, raw weights do not live on a natural grid.
Instead, locality is \emph{architectural}: weights incident to the same
neuron/channel are coupled; kernels share spatial structure; attention
blocks couple across layers through residual streams; and symmetries
introduce many equivalent coordinate representations.

We therefore view tokenization as choosing an \emph{inductive bias for
dependency structure}. 
Three families dominate practice: see next. We believe the practical trend is toward \emph{hybrid locality}: preserve intra-layer
structure via chunking and tags, while learning inter-layer dependencies
via sequence models, recurrence, or diffusion refinements.

\vspace{-0.7em}\paragraph{(1) Flattening and chunking (sequence view).}
A naive approach flattens all parameters into one long vector, but this
destroys layer structure and aggravates symmetry.
Early hypernetwork-based NAS (e.g., SMASH~\cite{brock2018smash})
used coarse flattening, while subsequent work emphasized
structured partitions and function-preserving transformations such as
Net2Net~\cite{chen2016net2net}.

Modern large-scale generators partition weights into coherent tokens.
RPG~\cite{wang2024rpg} slices each layer into fixed-length chunks, applies per-layer
normalization, and appends positional tags (layer index, chunk index) to
break permutation ambiguity across tokens.
SANE similarly operates on thousands of tokens, but encodes them into
learned embeddings rather than directly modeling raw floats
\cite{schurholt2024towards}.
This ``token-with-position'' paradigm implements a weak form of locality:
it assumes that chunks within a layer are more correlated than chunks
across layers, while still permitting the generator to learn global
dependencies through attention or recurrence.

\vspace{-0.7em}\paragraph{(2) Set encodings (permutation-invariant view).}
Permutation symmetry (\$~\ref{sec:geometry}) suggests representing
collections of neurons/channels as \emph{unordered sets}.
Set encoders and permutation-invariant pooling attempt to quotient out
label symmetries at the representation level.
For example, Set-Based Neural Network Encoding~\cite{andreis2023set}
treats layer parameters as sets of neuron vectors and uses
permutation-invariant operations to form representations.
Such tokenization can reduce spurious variation, but may
fail to model long-range inter-layer correlations.

A practical lesson from \textit{Git Re-Basin}~\cite{ainsworth2022git} and
related alignment work is that \emph{symmetry can be handled either
before modeling (canonicalization / alignment) or inside the model
(permutation-invariant encoders)}.
Most current generators choose the latter implicitly (via tags,
invariances, and data augmentation), because perfect canonicalization is
itself a hard problem.

\vspace{-0.7em}\paragraph{(3) Graph tokenization (architecture-as-graph view).}
When architectures vary, tokenization naturally becomes graph-structured.
Graph HyperNetworks~\cite{knyazev2021parameter} treat the network as a DAG
$\mathcal{G}=(\mathcal{V},\mathcal{E})$, with node attributes encoding
layer types and shapes; message passing propagates context and produces a
weight prediction per node.
Graph MetaNetworks~\cite{lim2023graph} extend this idea to diverse
architectures, aiming to predict parameters for unseen graphs.
Graph tokenization is the clearest path toward regime (iii), but its
ability to scale to modern foundation models is still limited by memory
and by the difficulty of capturing dense long-range dependencies.

%---------------------------------------------------------------------\
\vspace{-0.5em}
\subsection{Embedding and Latent Representation Design}
\label{sec:embedding}

Embedding compresses tokenized weights (and conditions) into latent
variables that make generation tractable.
This stage operationalizes two geometric facts from
Section~\ref{sec:geometry}:
(i) good solutions lie on a thin manifold, and
(ii) the density on that manifold is highly non-uniform.

\vspace{-0.7em}\paragraph{Autoencoding and layer-wise balancing.}
Early work showed that simple weight statistics predict performance
\cite{unterthiner2020predicting}.
Hyper-Representations~\cite{schurholt2022hyper} learn an autoencoder over
entire networks, and emphasize \emph{layer-wise loss normalization} so
small layers are not ignored.
SANE~\cite{schurholt2024towards} produces layer-wise latents from large
token sequences, enabling scalable encoding of large networks. Later, diffusion-style approaches often use an explicit encoder-decoder pair
$E:\mathbb{R}^{d}\!\to\!\mathbb{R}^{d'}$ with $d'\!\ll d$ to run the
generative model in latent space.
\cite{wang2024neural} compresses a $\sim$100M ConvNeXt into a
much smaller latent and performs diffusion in that space, mitigating the curse of dimensionality.

It has been noted that well-structured latents often admit semantic arithmetic:
interpolation or vector arithmetic can produce predictable changes in
behavior~\cite{schurholt2022hyper,li2024tina}.
It is a route to controllable generation
and compositional model editing.

\vspace{-0.7em}\paragraph{Conditioning: from prompts and datasets to weights.}
Conditioning is where ``prompt-to-model'' becomes concrete.
Recent systems increasingly treat conditions as \emph{task fingerprints}:
a few prompts, a dataset descriptor, an image identity embedding, or an
architecture graph.

D2NWG conditions diffusion on dataset descriptors to emit models for
multiple target domains~\cite{soro2024diffusion}.
Tina uses a CLIP text embedding of a user prompt to generate custom classifiers~\cite{li2024tina}.
HyperDreamBooth conditions on image embeddings to produce identity
adapters~\cite{ruiz2024hyperdreambooth}. A notable recent step is prompt-conditioned \emph{adapter generation} for
LLMs:
\cite{liang2025drag} learns a map
from a small batch of unlabeled task prompts to LoRA updates, enabling
task adaptation without per-task optimization.
This strengthens the argument that conditioning need not be a handcrafted
metadata vector; it can be directly extracted from raw task prompts.

%---------------------------------------------------------------------
\vspace{-0.5em}
\subsection{Generative Predictors for Weight Synthesis}
\label{sec:predictors}

We now summarize the predictor families and take an opinionated stance on \emph{where each is practically strongest}.

\vspace{-0.7em}\paragraph{Hypernetworks (fastest, best for personalization).}
Classical HyperNetworks generate weights from low-dimensional codes
\cite{ha2017hypernetworks}.
They remain the most pragmatic choice when latency matters and the
output is parameter-efficient (e.g., LoRA-like updates).
HyperDreamBooth exemplifies this: a lightweight hypernetwork generates
low-rank personalization updates in seconds~\cite{ruiz2024hyperdreambooth}.
In regime (i), hypernetworks currently offer the cleanest deployment
path.

\vspace{-0.7em}\paragraph{Graph-conditioned predictors (cross-architecture, but scaling-limited).}
GHN-style models predict weights from architecture graphs
\cite{knyazev2021parameter} and can generalize to unseen DAGs, enabling
``zero-shot'' evaluation in NAS-like workflows.
Graph MetaNetworks~\cite{lim2023graph} push toward broader architectural
coverage.
However, scaling these predictors to modern foundation models requires
memory-safe graph representations and stronger inductive bias for
long-range dependencies.

\vspace{-0.7em}\paragraph{Diffusion in weight space (expressive, currently best scaling evidence).}
Diffusion-based generators perturb real weights with noise and learn to
denoise.
D2NWG~\cite{soro2024diffusion} shows dataset-conditioned generation.
Neural Network Diffusion demonstrates high-quality generation for large
vision backbones in a compressed latent space~\cite{wang2024neural}.
More recent hybrid diffusion stacks (e.g., recurrent encoders plus
diffusion refinement) provide the strongest current evidence for
mid-scale full-weight generation.
We therefore view diffusion (particularly latent or hybrid diffusion) as
the most promising direction for regime (ii), where distributional
coverage and fidelity dominate.

\vspace{-0.7em}\paragraph{Normalizing flows (one-step sampling, promising but less mature).}
Flow-matching approaches such as FLoWN offer invertible, one-step
sampling of weights~\cite{saragih2025flow}, with potential advantages in
controllability and likelihood-based evaluation.
Their maturity and scale are currently behind diffusion,
but they remain attractive where low-latency conditional sampling is
critical.

\vspace{-0.7em}\paragraph{Hybrid stacks.}
Hybrid predictors explicitly implement a curriculum:
first model global inter-layer structure, then refine local details.
RPG \cite{wang2024rpg} is emblematic: a recurrent proto-encoder captures long-range
dependencies, and diffusion fills token-level detail
\cite{wang2024rpg}.

We note that RPG represents a milestone by efficiently synthesizing entire ConvNeXts or ViTs (up to ~200M weights) in a single GPU pass - the first approach to go beyond toy 10M-parameter models to competitive 100M+ parameters, generated on commodity hardware. RPG-produced ConvNeXt-L matches conventional training within a few tenths of a percent on ImageNet. While other diffusion-based approaches \cite{wang2024neural,soro2024diffusion} addressed scaling through modular generation, they often missed inter-block correlations, yielding suboptimal results. 

%---------------------------------------------------------------------
\vspace{-0.5em}
\subsection{Training Strategies and Collapse Avoidance}
\label{sec:training}

Training weight generators differs from standard generative modeling in
two ways: (i) the data distribution is highly multi-modal (many tasks,
architectures, and training recipes), and (ii) naive maximum-likelihood
can collapse to memorization of a few basins.
Practical training therefore relies on various stabilization techniques.

\vspace{-0.7em}\paragraph{Augmentation and regularization.}
Basic noise injection and dropout over tokens mitigate memorization.
MixUp-in-weight-space enlarges distributional support and improves
generalization~\cite{shamsian2024augmentations}.

Contrastive objectives can encourage invariance to permutations and
training artifacts~\cite{schurholt2021self}.
For INR generators, rendering-based fidelity losses (e.g., IoU penalties)
provide functional supervision beyond raw weight error
\cite{erkoc2023hyperdiffusion}.

\vspace{-0.7em}\paragraph{Curriculum learning.} 
Practical curricula include shorter diffusion chains, smaller token
sets, and progressively increasing context length, as in RPG
\cite{wang2024rpg}.
Conceptually, this corresponds to learning global structure first (easy)
and token-level detail later (hard), matching the hierarchical
dependence structure suggested by Section~\ref{sec:geometry}.

\vspace{-0.7em}\paragraph{Diverse model zoos and task coverage.}
Diverse corpora reduce collapse by forcing the generator to explain
multiple basins.
D2NWG explicitly trains on heterogeneous checkpoint zoos
\cite{soro2024diffusion}.
\cite{peebles2022learning} scale supervision dramatically, training on
tens of millions of checkpoints to learn generative update rules.
SANE aggregates multiple public model zoos and achieves strong zero-shot
viability without overfitting~\cite{schurholt2024towards}.

%---------------------------------------------------------------------
\vspace{-0.5em}
\subsection{Evaluation Metrics: Beyond One Simple Accuracy}
\label{sec:metrics}

A core critique of early weight-generation papers is that they over-index
on task accuracy and under-specify how to evaluate generalization,
novelty, and reliability.
Evaluation must therefore be multi-axis, reflecting the fact that many
distinct weights can yield similar accuracy but differ sharply in
robustness, calibration, privacy risk, or mergeability.

\vspace{-0.7em}\paragraph{Three-Axis Framework.} We believe that any viable benchmark should report three axes as naturally induced by Fig. \ref{fig:achievements_and_promise}: (A) \textit{within-architecture generation}, where we must test against nearest-neighbor retrieval, aligned interpolation, and model soups; (B) \textit{conditional adaptation}, where prompt-, dataset-, or user-conditioned generators are compared against PEFT/fine-tuning under matched wall-clock and memory budgets; and (C) \textit{cross-architecture generation}, where graph-conditioned generators are evaluated on held-out architecture families and unseen tensor schemas. Then, along each axis, evaluation should include downstream performance, generation cost, diversity and ensemble gain, calibration, robustness, memorization, provenance, and safety.

\vspace{-0.7em}\paragraph{Task performance and adaptation efficiency.}
Accuracy (or BLEU / RL return) remains primary, but should be reported
alongside adaptation cost: GPU-hours, wall-clock generation time, and
memory footprint.
For example, RPG-generated ConvNeXt-L matches fully trained ImageNet
accuracy in under 90 seconds~\cite{wang2024rpg}.
For personalization, \emph{time-to-personalize} is often the key metric.

\vspace{-0.7em}\paragraph{Novelty vs. interpolation baselines.}
A generator is considered novel only if it outperforms strong
baselines such as aligned interpolation \cite{ainsworth2022git} or model soups \cite{wortsman2022model} at comparable
distance. This concern is not hypothetical: recent diagnostic work reports that
several representative weight generators produce replicas or simple
interpolations of training checkpoints, and fail to outperform simple
noise or ensemble baselines when novelty and performance are required jointly
\cite{zeng2025generative}. Representation similarity measures such as CKA~\cite{kornblith2019similarity}
help detect trivial memorization, while spectral signatures (e.g., singular
value distributions) test whether the generator matches the geometry of trained solutions.

\vspace{-0.7em}\paragraph{Diversity and mode coverage.}
Since many solutions exist, generators should be evaluated for ensemble
utility and mode coverage.
Pairwise disagreement, ensemble gains, and performance variance across
samples quantify whether the generator populates multiple basins.

\vspace{-0.7em}\paragraph{Robustness, calibration, and trust metrics.}
Even within connected basins, models differ in robustness and
calibration.
Standard tools include NLL under corruption, ECE, and adversarial or
stress-test evaluation.
Model soups demonstrate that weight averaging can improve robustness
without extra inference cost~\cite{wortsman2022model}, motivating
evaluation against such strong non-generative baselines.
Depending on deployment, evaluation should extend to fairness and safety
benchmarks such as trust and privacy assessment suites
\cite{wang2023decodingtrust,li2024llm}.

\vspace{-0.5em}\section{Applications Enabled by Weight Generators}
\label{sec:applications}

\subsection{Instant Personalization} \vspace{-0.2em}

Weight generation has revolutionized model personalization across domains by
amortizing adaptation into a single forward pass.
In visual generation, HyperDreamBooth~\cite{ruiz2024hyperdreambooth}
dramatically accelerates subject-specific adaptation by producing weight
updates in one pass, achieving personalization in $\sim$20 seconds
($25\times$ faster than DreamBooth~\cite{ruiz2023dreambooth}) while
preserving base-model knowledge and style diversity.

Beyond images, Tina~\cite{li2024tina} enables \emph{text-to-model}
generation, instantiating task-specific classifiers directly from textual
descriptions using a single diffusion model.
D2NWG~\cite{soro2024diffusion} further generalizes this paradigm to
dataset-conditioned full weight synthesis, matching standard transfer
learning accuracy while improving few-shot performance by $\sim$6\%, and
extending to language models where it enhances LLaMA’s mathematical
reasoning.
Most recently, \cite{liang2025drag} enable on-the-fly task adaptation of LLMs: a user simply provides a description of a new task (e.g. a specific style of Q\&A or a new programming API), and the system drops in a generated LoRA adapter that immediately makes the LLM perform well on that task. They demonstrated strong zero-shot generalization to tasks unseen during generator training, such as reasoning puzzles and multimodal queries, with performance even above that of standard fine-tuning.

\vspace{-0.5em}
\subsection{Model Fusion and Editing} \vspace{-0.2em}

Generative weight models provide new tools for combining knowledge from multiple models or domains. 
Traditional model fusion techniques typically require models to be trained on related data and carefully aligned in a shared coordinate frame, and even then produce only crude combinations that may underperform on complex tasks.
In contrast, generative approaches can learn a weight-space distribution or \emph{manifold} that encompasses diverse domains, enabling smooth interpolation or conditional sampling.

For example, \cite{peebles2022learning} showed that training a generative model on checkpoints from both image classification and language tasks yields latent interpolations that produce models with mixed characteristics, hinting at cross-domain fusion through learned manifolds.
Recent work \cite{dravid2024interpreting} on interpreting the weight space of customized diffusion models demonstrates this phenomenon at scale: by populating a large dataset of fine-tuned diffusion checkpoints and projecting them into a low-dimensional subspace, they found that navigating within this learned subspace can generate new diffusion models encoding novel identities, and that linear directions in this space correspond to semantic edits (e.g., adding visual attributes) in the resulting models.
This empirical evidence suggests that fine-tuned model weights behave as an interpretable meta-latent space where interpolation and controlled traversal produce meaningful, high-performing models.

\vspace{-0.5em}
\subsection{Efficient Neural Architecture Search}\vspace{-0.2em}

Weight generation transforms Neural Architecture Search (NAS) by eliminating costly training of candidate architectures. By training hypernetworks that map architectures to performant weights \cite{brock2018smash}, researchers can instantly predict weights for unseen architectures. 

GHN \cite{knyazev2021parameter}, trained on one million neural nets, can predict on unseen and diverse networks, including all 24 million weights of a ResNet-50 in one forward pass, achieving 50\% ImageNet top-5 accuracy and 60\% CIFAR-10 top-1 without any SGD. These predicted weights serve as excellent initializations that can be quickly fine-tuned. Recent advances like GHN-3 \cite{knyazev2023can} use Transformer-based hypernetworks to predict weights that outperform standard one-epoch SGD training, making ``zero-shot'' NAS feasible for exploring ultra-large design spaces.

\vspace{-0.5em}
\subsection{On-Device Learning} \vspace{-0.2em}

Weight-generation methods also have important implications for on-device learning. By performing learning in a lightweight hypernetwork and deploying only its generated weights, the device does not undergo heavy training, and sensitive user data never leave the device. Several systems demonstrate its effectiveness. SecDOOD~\cite{li2025secure} employs a cloud-based hypernetwork for video OOD detection on IoT devices, mapping encrypted feature summaries to specialized model weights without exposing raw data. In federated learning, pFedHN~\cite{shamsian2021personalized} utilizes a global hypernetwork to generate personalized models for each client, sharing only hypernetwork parameters while significantly outperforming traditional federated averaging and generalizing better to new clients.

\vspace{-0.5em}
\section{The Next Frontier: Scaling Up Weight Generation to Foundation Models?}

While synthesizing arbitrary trillion-parameter models end-to-end remains an open challenge, conditional weight generation has recently scaled far beyond toy settings and is advancing rapidly. It now operates on nontrivial large backbones, where deployment cost, adaptation latency, and repeated fine-tuning overhead are major practical bottlenecks.

The clearest evidence to date is \textbf{HY-WU}~\cite{tencent2026hy}, which
recasts conditional weight generation as a \emph{functional neural
memory}: instead of repeatedly overwriting a shared parameter vector, a
generator synthesizes instance-conditioned LoRA updates on the fly and
injects them into a frozen backbone without test-time optimization.
Notably, HY-WU is demonstrated on an 80B multimodal FM backbone with
13B active parameters, using an 8.11B generator to produce 0.72B LoRA
parameters.
This matters because it moves weight generation out of the ``small-model curiosity''
regime and into a setting where conditional weight generation is already
interfacing with genuinely large-scale foundation-model infrastructure in a leading industry lab.

This evidence is reinforced by a broader scaling trend.
RPG~\cite{wang2024rpg} generates competitive ConvNeXt and ViT backbones
with up to $\sim$200M parameters by explicitly modeling inter-layer
dependencies.
Text-to-LoRA \cite{charakorn2025text} generates LoRA updates for Llama-3.1-8B and Gemma-2-2B;
Doc-to-LoRA \cite{charakorn2026doc} does so for Gemma-2-2B, Mistral-7B, and Qwen3-4B; and SHINE \cite{liu2026shine}
extends the regime to Qwen3-8B.
These results suggest a consistent staged picture:
today's strongest evidence is not unrestricted full-checkpoint
generation at frontier scale, but large-backbone, adapter-scale, and
conditional weight generation.

At the same time, the \textbf{main barriers} to scaling are not merely matters of raw compute, but of structure. Representation alignment, long-range dependencies, and checkpoint heterogeneity are fundamental technical challenges rather than secondary engineering details. These factors will likely determine whether the current adapter-scale successes can eventually extend to broader full/base model-generation regimes. We discuss these issues below:

\vspace{-0.7em}
\paragraph{i. Representation alignment \& symmetry.}
Raw weights are not canonical.
Permutation and scaling symmetries create many numerically distinct
parameterizations of the same function.
Without alignment, quotient-aware encoders, or symmetry, a
generator may waste capacity modeling representation artifacts rather
than functional variation.

\vspace{-0.7em}\paragraph{ii. Long-range dependency and memory scaling.}
Large models contain cross-layer and cross-module dependencies that are
poorly captured by naive flattening or purely local chunking.
Scaling likely requires (more) hierarchical tokenization, latent or recurrent
generators, sparse graph attention, and scoped generation of parameter
subsets rather than monolithic modeling of all tensors at once.

\vspace{-0.7em}\paragraph{iii. Architecture and training heterogeneity.}
Checkpoints differ by architecture, tokenizer, preprocessing,
optimizer, scheduler, regularization, objective, data mixture, and
fine-tuning recipe.
Treating all checkpoints as samples from one unconditioned density is
unlikely to succeed.
A more realistic target is the conditional family $
p(W \mid A, C, R)$ (Fig. \ref{fig:achievements_and_promise}).
In practice, such metadata should be used for filtering,
stratification, or conditioning, rather than ignored.

\vspace{-0.7em}\paragraph{iv. Memorization, provenance, and safety.}
A generator trained on checkpoint zoos can inherit proprietary weights,
backdoors, biases, or contamination.
Weight generation will therefore face a ``model-supply-chain" problem:
provenance, lineage tracking, memorization tests, safety audits, and
quarantine of untrusted checkpoints.

\vspace{-0.5em}\section{Critical Discussions}\vspace{-0.2em}
\label{sec:limitations}

We clarify what weight-space generation \emph{can} and \emph{cannot}
currently do, and confront credible counter-positions.

\vspace{-0.7em}\subsection{Can Generation Go Beyond Optimizer Solutions?}\vspace{-0.2em}

A central question is whether weight generators merely reproduce the
outcomes of SGD or can synthesize \emph{novel} solutions.
At present, most empirical results show parity with well-tuned
optimization rather than consistent superiority.
However, novelty should be evaluated structurally rather than
numerically.
Generative models are trained across many checkpoints and tasks, and
thus have access to cross-model correlations unavailable to single-run
optimization.
Evidence from generative checkpoint models~\cite{peebles2022learning}
and from interpretable diffusion subspaces~\cite{dravid2024interpreting}
shows that generators can sample distinct regions within the same
connected basin, differing in robustness, calibration, or semantic
attributes.
Whether such diversity can be systematically exploited to outperform
SGD on core metrics remains an open question, but the mechanism,
namely sampling from a learned density rather than following a single
trajectory, is fundamentally different. Recent analyses caution that current generators
do not automatically achieve such distributional novelty \citep{zeng2025generative}.

\vspace{-0.7em}\subsection{The Limits of Weight Space Learnability}\vspace{-0.2em}

Another concern is whether weight space exhibits sufficient locality or
regularity to be learnable at scale. We make clear that locality in weight
space is \emph{architectural}, not spatial. Dependencies are long-range,
module-based, and symmetry-entangled. This is why naive token-wise
models fail and why structured tokenizers, hierarchical generators, and
curriculum learning are necessary. Modularity offers promise for
compositional generalization, but boundaries between modules are neither
smooth nor universal. Hence, generation is feasible but non-trivial:
it requires inductive biases that mirror those of successful optimization. Without sufficient checkpoint diversity, and held-out tests against retrieval, 
interpolation, or perturbation baselines, a generator may learn a
compressed description of its training zoo rather than a reusable model
distribution \citep{zeng2025generative}.

\vspace{-0.4em}\section{Alternative Views}
\label{sec:altviews}

\vspace{-0.3em}
\paragraph{$\rhd$ Alternative View 1:} \textit{``Fine-tuning, PEFT, Transfer Learning, and In-Context Learning are Sufficient.''}
One might argue that PEFT, LoRA, transfer learning, and in-context
learning already provide fast and practical adaptation.
We agree that these paradigms remain attractive when
iterative per-task optimization is acceptable, task data are available,
and strong task-specific adaptation is desired.
In-context learning is preferable when no persistent parameter change is
needed.
Our response is therefore economic rather than purely algorithmic:
weight generation targets settings where the cost of repeated
optimization or long context windows is itself the bottleneck, for
example in low-latency or privacy-constrained deployment.
Its goal is to amortize that cost across tasks and users, enabling
second-scale adaptation when deployment-time optimization is
undesirable.

Continual learning is another adjacent but distinct paradigm. It focuses
on updating one deployed model over time while avoiding forgetting. By
contrast, weight generation aims to learn a distribution over reusable
parameter updates or checkpoints that can be sampled without committing
all knowledge to a single evolving parameter vector. It does not
solve forgetting by itself, but it could
complement continual learning by generating task- or episode-specific
updates while keeping provenance and rollback explicit.

\vspace{-1em}
\paragraph{$\rhd$ Alternative View 2:} \textit{``Yet Another Meta-Learning or Model Merging Under a New Name.''}
Weight generation overlaps with both meta-learning and model
merging, but the primitive it studies is different.
Meta-learning typically operates over tasks with shared structure and
still culminates in iterative adaptation.
Model merging is closer in spirit, but usually assumes a small set of
compatible source models and simple algebraic composition.
By contrast, weight generation treats \emph{trained models
themselves} as data and asks whether useful parameter updates can be
produced in a single feed-forward pass from conditioning information.
This distinction matters for evaluation, infrastructure, and
governance.
We do not claim that weight-space generation already dominates these
approaches, but rather that it opens a distinct amortized one-shot
regime that remains under-standardized relative to its promise.

\vspace{-1.1em}
\paragraph{$\rhd$ Alternative View 3:} \textit{``Weight  Generation Will Not Matter Until It Outperforms Gradient-Based Optimization End-to-End for Full Foundation-Model Training.''}
We do not believe this is the right threshold.
A new primitive can become foundational without immediately replacing
the entire optimization stack. It may first prove its value by replacing
or amortizing expensive subroutines that recur throughout modern ML
pipelines.
In our view, the strongest current case for weight generation lies
precisely in this regime: adaptation, personalization, context
internalization, and model-space search, where current practice
repeatedly invokes optimization to produce relatively small but
operationally valuable weight updates. From this perspective, frontier-scale full-model synthesis remains an
important open research problem rather than the central bar that the
field must clear immediately. 
At the same time, as discussed in \S 4.5, there is growing evidence that
the field is making meaningful progress toward larger-scale regimes.

\vspace{-0.6em}
\section{Conclusion and Call to Action}
\vspace{-0.3em}
\label{sec:conclusion}

This paper argues that neural network checkpoints should be treated not only
as optimization outputs, but as a first-class data modality for generative
modeling. We believe the path forward is not a single universal weight generator, but a
standardized research stack. \textit{First}, the community needs checkpoint-as-data
infrastructure: curated model zoos with architecture schemas, training recipes,
quality traces, provenance, licensing, and dataset lineage. \textit{Second}, WSG methods
must respect the structure of weight space, including permutation symmetry,
low-rank and modular structure, optimizer-induced bias, and long-range
cross-layer dependencies. \textit{Third}, evaluation must move beyond task accuracy to
multi-axis benchmarks that measure efficiency, novelty beyond retrieval or
interpolation, diversity, robustness, calibration, memorization, and safety.
\textit{Fourth}, public weight generators should include provenance tracking,
watermarking, privacy audits, and safety screening as part of the release
protocol rather than as afterthoughts. If these standards are built, weight-space generation can turn repeated
artisanal optimization into a reusable and auditable model-production process. 
The field should advance through deliberate infrastructure, benchmark, and governance design.

\section*{Impact Statement}

This paper advocates treating neural network checkpoints as a first-class
data modality and standardizing generative modeling in weight space as a
core machine learning primitive.
If adopted, this paradigm could substantially reduce the computational
and environmental cost of repeated model training, enable rapid and
privacy-preserving personalization, and lower barriers for deploying
specialized models on edge and resource-constrained devices.

At the same time, generative access to model weights raises legitimate
concerns around memorization, misuse, and model
provenance. 
A generator may memorize proprietary weights, raising IP and copyright
concerns analogous to data leakage in LLMs~\cite{li2024llm}.
Automated backdoor insertion or mass production of harmful models are
also plausible misuse scenarios.
Mitigations will require provenance tracking, watermarking of generated
weights, controlled access to training corpora, and evaluation protocols
that test for memorization and malicious behavior.
These concerns strengthen the case for rigorous community
standards. Overall, the anticipated societal impact is dual-use: weight-space
generation has the potential to democratize and accelerate AI
deployment, but only if accompanied by appropriate governance and
safeguards.

\bibliographystyle{icml2026}
\bibliography{weightgen}

@inproceedings{ha2017hypernetworks,
  title = {HyperNetworks},
  author = {Ha, David and Dai, Andrew M. and Le, Quoc V.},
  booktitle = {ICLR},
  year = {2017}
}

@inproceedings{brock2018smash,
title={{SMASH}: One-Shot Model Architecture Search through HyperNetworks},
author={Andrew Brock and Theo Lim and J.M. Ritchie and Nick Weston},
booktitle={ICLR},
year={2018}
}

@article{goodfellow2014qualitatively,
  title={Qualitatively characterizing neural network optimization problems},
  author={Goodfellow, Ian J and Vinyals, Oriol and Saxe, Andrew M},
  journal={ICLR},
  year={2015}
}

@inproceedings{song2025does,
  title={Does SGD really happen in tiny subspaces?},
  author={Song, Minhak and Ahn, Kwangjun and Yun, Chulhee},
  booktitle={ICLR},
  year={2025}
}

@inproceedings{choromanska2015loss,
  title={The loss surfaces of multilayer networks},
  author={Choromanska, Anna and Henaff, Mikael and Mathieu, Michael and Arous, G{\'e}rard Ben and LeCun, Yann},
  booktitle={AISTATS},
  year={2015}
}

@inproceedings{knyazev2021parameter,
  title = {Parameter Prediction for Unseen Deep Architectures},
  author = {Knyazev, Boris and Drozdzal, Michal and Taylor, Graham W. and Romero-Soriano, Adriana},
  booktitle = {NeurIPS},
  year = {2021}
}

@article{unterthiner2020predicting,
  title = {Predicting Neural Network Accuracy from Weights},
  author = {Unterthiner, Thomas and Keysers, Daniel and Gelly, Sylvain and Bousquet, Olivier and Tolstikhin, Ilya},
  journal = {arXiv},
  year = {2020}
}

@article{peebles2022learning,
  title = {Learning to Learn with Generative Models of Neural Network Checkpoints},
  author = {Peebles, William S. and Radosavovic, Ilija and Brooks, Tim and Efros, Alexei A. and Malik, Jitendra},
  journal = {arXiv},
  year = {2022}
}

@article{wang2024neural,
  title = {Neural Network Diffusion},
  author = {Wang, Kai and Tang, Dongwen and Zeng, Boya and Yin, Yida and Xu, Zhaopan and Zhou, Yukun and Zang, Zelin and Darrell, Trevor and Liu, Zhuang and You, Yang},
  journal = {arXiv},
  year = {2024}
}

@inproceedings{jiang2019fantastic,
title={Fantastic Generalization Measures and Where to Find Them},
author={Yiding Jiang and Behnam Neyshabur and Hossein Mobahi and Dilip Krishnan and Samy Bengio},
booktitle={ICLR},
year={2020}
}

@inproceedings{zhu2024apollo,
  title={Apollo: Sgd-like memory, adamw-level performance},
  author={Zhu, Hanqing and Zhang, Zhenyu and Cong, Wenyan and Liu, Xi and Park, Sem and Chandra, Vikas and Long, Bo and Pan, David Z and Wang, Zhangyang and Lee, Jinwon},
  booktitle={MLSys},
  year={2025}
}

@article{mao2024training,
  title={The training process of many deep networks explores the same low-dimensional manifold},
  author={Mao, Jialin and Griniasty, Itay and Teoh, Han Kheng and Ramesh, Rahul and Yang, Rubing and Transtrum, Mark K and Sethna, James P and Chaudhari, Pratik},
  journal={PNAS},
  year={2024}
}

@inproceedings{zhao2024galore,
  title={Galore: Memory-efficient llm training by gradient low-rank projection},
  author={Zhao, Jiawei and Zhang, Zhenyu and Chen, Beidi and Wang, Zhangyang and Anandkumar, Anima and Tian, Yuandong},
  booktitle={ICML},
  year={2024}
}

@inproceedings{
frankle2018lottery,
title={The Lottery Ticket Hypothesis: Finding Sparse, Trainable Neural Networks},
author={Jonathan Frankle and Michael Carbin},
booktitle={ICLR},
year={2019}
}

@inproceedings{chen2020lottery,
  title={The lottery ticket hypothesis for pre-trained bert networks},
  author={Chen, Tianlong and Frankle, Jonathan and Chang, Shiyu and Liu, Sijia and Zhang, Yang and Wang, Zhangyang and Carbin, Michael},
  booktitle={NeurIPS},
  year={2020}
}

@inproceedings{zhao2024texttt,
  title={Model-GLUE: Democratized LLM Scaling for A Large Model Zoo in the Wild},
  author={Zhao, Xinyu and Sun, Guoheng and Cai, Ruisi and Zhou, Yukun and Li, Pingzhi and Wang, Peihao and Tan, Bowen and He, Yexiao and Chen, Li and Liang, Yi and others},
  booktitle={NeurIPS},
  year={2024}
}

@inproceedings{
entezari2021role,
title={The Role of Permutation Invariance in Linear Mode Connectivity of Neural Networks},
author={Rahim Entezari and Hanie Sedghi and Olga Saukh and Behnam Neyshabur},
booktitle={ICLR},
year={2022}
}

@inproceedings{petzka2021relative,
  title={Relative flatness and generalization},
  author={Petzka, Henning and Kamp, Michael and Adilova, Linara and Sminchisescu, Cristian and Boley, Mario},
  booktitle={NeurIPS},
  year={2021}
}

@inproceedings{huh2024platonic,
  title={The Platonic Representation Hypothesis},
  author={Huh, Minyoung and Cheung, Brian and Wang, Tongzhou and Isola, Phillip},
  booktitle={ICML},
  year={2024}
}

@inproceedings{schurholt2022hyper,
  title={Hyper-representations as generative models: Sampling unseen neural network weights},
  author={Sch{\"u}rholt, Konstantin and Knyazev, Boris and Gir{\'o}-i-Nieto, Xavier and Borth, Damian},
  booktitle={NeurIPS},
  year={2022}
}

@inproceedings{andreis2023set,
  title={Set-based neural network encoding without weight tying},
  author={Andreis, Bruno and Soro, Bedionita and Torr, Philip and Hwang, Sung Ju},
  booktitle={NeurIPS},
  year={2024}
}

@inproceedings{erkoc2023hyperdiffusion,
  title={{HyperDiffusion: Generating Implicit Neural Fields with Weight-Space Diffusion}},
  author={Erko\c{c}, Ziya and Ma, Fangchang and Shan, Qi and Nie{\ss}ner, Matthias and Dai, Angela},
  booktitle={ICCV},
  year={2023}
}

@inproceedings{chen2016net2net,
  title={Net2Net: Accelerating Learning via Knowledge Transfer},
  author={Chen, Tianqi and Goodfellow, Ian and Shlens, Jonathon},
  booktitle={ICLR},
  year={2016}
}

@inproceedings{gong2019efficient,
  title={Efficient training of bert by progressively stacking},
  author={Gong, Linyuan and He, Di and Li, Zhuohan and Qin, Tao and Wang, Liwei and Liu, Tieyan},
  booktitle={ICML},
  year={2019}
}

@inproceedings{wang2023learning,
title={Learning to grow pretrained models for efficient transformer training},
author={Wang, Peihao and Panda, Rameswar and Hennigen, Lucas Torroba and Greengard, Philip and Karlinsky, Leonid and Feris, Rogerio and Cox, David Daniel and Wang, Zhangyang and Kim, Yoon},
booktitle={ICLR},
year={2023}
}

@inproceedings{wang2023data,
  title={Data efficient neural scaling law via model reusing},
  author={Wang, Peihao and Panda, Rameswar and Wang, Zhangyang},
  booktitle={ICML},
  year={2023}
}

@inproceedings{yang2022deep,
  title={Deep model reassembly},
  author={Yang, Xingyi and Zhou, Daquan and Liu, Songhua and Ye, Jingwen and Wang, Xinchao},
  booktitle={NeurIPS},
  year={2022}
}

@article{wang2025neural,
  title={Why Neural Network Can Discover Symbolic Structures with Gradient-based Training: An Algebraic and Geometric Foundation for Neurosymbolic Reasoning},
  author={Wang, Peihao and Wang, Zhangyang},
  journal={arXiv},
  year={2025}
}

@article{kaushik2025universal,
  title={The universal weight subspace hypothesis},
  author={Kaushik, Prakhar and Chaudhari, Shravan and Vaidya, Ankit and Chellappa, Rama and Yuille, Alan},
  journal={arXiv},
  year={2025}
}

@inproceedings{liang2025drag,
  title={Drag-and-Drop LLMs: Zero-Shot Prompt-to-Weights},
  author={Liang, Zhiyuan and Tang, Dongwen and Zhou, Yuhao and Zhao, Xuanlei and Shi, Mingjia and Zhao, Wangbo and Li, Zekai and Wang, Peihao and Sch{\"u}rholt, Konstantin and Borth, Damian and others},
  booktitle={NeurIPS},
  year={2025}
}

@inproceedings{wortsman2022model,
  title={Model soups: averaging weights of multiple fine-tuned models improves accuracy without increasing inference time},
  author={Wortsman, Mitchell and Ilharco, Gabriel and Gadre, Samir Ya and Roelofs, Rebecca and Gontijo-Lopes, Raphael and Morcos, Ari S and Namkoong, Hongseok and Farhadi, Ali and Carmon, Yair and Kornblith, Simon and others},
  booktitle={ICML},
  year={2022}
}

@inproceedings{wang2023decodingtrust,
  title={DecodingTrust: A Comprehensive Assessment of Trustworthiness in GPT Models.},
  author={Wang, Boxin and Chen, Weixin and Pei, Hengzhi and Xie, Chulin and Kang, Mintong and Zhang, Chenhui and Xu, Chejian and Xiong, Zidi and Dutta, Ritik and Schaeffer, Rylan and others},
  booktitle={NeurIPS},
  year={2023}
}

@inproceedings{li2024llm,
  title={Llm-pbe: Assessing data privacy in large language models},
  author={Li, Qinbin and Hong, Junyuan and Xie, Chulin and Tan, Jeffrey and Xin, Rachel and Hou, Junyi and Yin, Xavier and Wang, Zhun and Hendrycks, Dan and Wang, Zhangyang and others},
  booktitle={VLDB},
  year={2024}
}

@inproceedings{
saragih2025flow,
title={Flow to Learn: Flow Matching on Neural Network Parameters},
author={Daniel Saragih and Deyu Cao and Tejas Balaji and Ashwin Santhosh},
booktitle={Workshop on Neural Network Weights as a New Data Modality},
year={2025}
}

@article{patterson2021carbon,
  title={Carbon emissions and large neural network training},
  author={Patterson, David and Gonzalez, Joseph and Le, Quoc and Liang, Chen and Munguia, Lluis-Miquel and Rothchild, Daniel and So, David and Texier, Maud and Dean, Jeff},
  journal={arXiv},
  year={2021}
}

@inproceedings{ruiz2023dreambooth,
  title={Dreambooth: Fine tuning text-to-image diffusion models for subject-driven generation},
  author={Ruiz, Nataniel and Li, Yuanzhen and Jampani, Varun and Pritch, Yael and Rubinstein, Michael and Aberman, Kfir},
  booktitle={CVPR},
  year={2023}
}

@inproceedings{wang2024rpg,
  title={Recurrent diffusion for large-scale parameter generation},
  author={Wang, Kai and Tang, Dongwen and Zhao, Wangbo and Sch{\"u}rholt, Konstantin and Wang, Zhangyang and You, Yang},
  booktitle={NeurIPS},
  year={2025}
}

@inproceedings{
lim2023graph,
title={Graph Metanetworks for Processing Diverse Neural Architectures},
author={Derek Lim and Haggai Maron and Marc T. Law and Jonathan Lorraine and James Lucas},
booktitle={ICLR},
year={2024}
}

@article{li2024tina,
  title={Text-to-model: Text-conditioned neural network diffusion for train-once-for-all personalization},
  author={Li, Zexi and Gao, Lingzhi and Wu, Chao},
  journal={arXiv},
  year={2024}
}

@inproceedings{ruiz2024hyperdreambooth,
  title={{HyperDreamBooth: HyperNetworks for Fast Personalization of Text-to-Image Models}},
  author={Ruiz, Nataniel and Li, Yuanzhen and Jampani, Varun and Wei, Wei and Hou, Tingbo and Pritch, Yael and Wadhwa, Neal and Rubinstein, Michael and Aberman, Kfir},
  booktitle={CVPR},
  year={2024}
}

@inproceedings{schurholt2021self,
  title={{Self-Supervised Representation Learning on Neural Network Weights for Model Characteristic Prediction}},
  author={Sch{\"u}rholt, Konstantin and Kostadinov, Dimche and Borth, Damian},
  booktitle={NeurIPS Workshop},
  year={2021}
}

@article{tencent2026hy,
  title={HY-WU (Part I): An Extensible Functional Neural Memory Framework and An Instantiation in Text-Guided Image Editing},
  author={Tencent HY Team and others},
  journal={arXiv},
  year={2026}
}

@inproceedings{liu2026shine,
  title={SHINE: A Scalable In-Context Hypernetwork for Mapping Context to LoRA in a Single Pass},
  author={Liu, Yewei and Wang, Xiyuan and Mao, Yansheng and Gelbery, Yoav and Maron, Haggai and Zhang, Muhan},
  booktitle={ICML},
  year={2026}
}

@inproceedings{charakorn2025text,
  title={Text-to-LoRA: Instant Transformer Adaption},
  author={Charakorn, Rujikorn and Cetin, Edoardo and Tang, Yujin and Lange, Robert Tjarko},
  booktitle={ICML},
  year={2025}
}

@article{charakorn2026doc,
  title={Doc-to-lora: Learning to instantly internalize contexts},
  author={Charakorn, Rujikorn and Cetin, Edoardo and Uesaka, Shinnosuke and Lange, Robert Tjarko},
  journal={arXiv},
  year={2026}
}

@inproceedings{jaiswal2025,
  title={From Low Rank Gradient Subspace Stabilization to Low-Rank Weights: Observations, Theories, and Applications},
  author={Jaiswal, Ajay Kumar and Wang, Yifan and Yin, Lu and Liu, Shiwei and Chen, Runjin and Zhao, Jiawei and Grama, Ananth and Tian, Yuandong and Wang, Zhangyang},
  booktitle={ICML},
  year={2025}
}

@inproceedings{zeng2025generative,
  title={Generative Modeling of Weights: Generalization or Memorization?},
  author={Boya Zeng and Yida Yin and Zhiqiu Xu and Zhuang Liu},
  booktitle={CVPR},
  year={2026},
}

@inproceedings{schurholt2024towards,
  title = {Towards Scalable and Versatile Weight Space Learning},
  author = {Sch{\"u}rholt, Konstantin and Mahoney, Michael W and Borth, Damian},
  booktitle = {ICML},
  year = {2024}
}

@inproceedings{soro2024diffusion,
  title = {Diffusion-Based Neural Network Weights Generation},
  author = {Soro, Bedionita and Andreis, Bruno and Lee, Hayeon and Jeong, Wonyong and Chong, Song and Hutter, Frank and Hwang, Sung Ju},
  booktitle = {ICLR},
  year = {2025}
}

@inproceedings{garipov2018loss, title={Loss Surfaces, Mode Connectivity, and Fast Ensembling of DNNs}, author={Garipov, Timur and Izmailov, Pavel and Podoprikhin, Dmitrii and Vetrov, Dmitry P. and Wilson, Andrew Gordon}, booktitle={NeurIPS}, year={2018} }

@inproceedings{draxler2018essentially, title={Essentially No Barriers in Neural Network Energy Landscape}, author={Draxler, Felix and Veschgini, Kambis and Salmhofer, Manfred and Hamprecht, Fred A.}, booktitle={ICML}, year={2018} }

@inproceedings{kuditipudi2019explaining, title={Explaining Landscape Connectivity of Low-Cost Solutions for Multilayer Nets}, author={Kuditipudi, Rohith and Wang, Xiang and Lee, Holden and Zhang, Yi and Li, Zhiyuan and Hu, Wei and Arora, Sanjeev and Ge, Rong}, booktitle={NeurIPS}, year={2019} }

@article{sagun2018,
  title={Empirical analysis of the hessian of over-parametrized neural networks},
  author={Sagun, Levent and Evci, Utku and Guney, V Ugur and Dauphin, Yann and Bottou, Leon},
  journal={arXiv},
  year={2017}
}

@inproceedings{cranmer2020discovering,
  title={Discovering symbolic models from deep learning with inductive biases},
  author={Cranmer, Miles and Sanchez Gonzalez, Alvaro and Battaglia, Peter and Xu, Rui and Cranmer, Kyle and Spergel, David and Ho, Shirley},
  booktitle={NeurIPS},
  year={2020}
}

@inproceedings{shamsian2021personalized,
  title={Personalized federated learning using hypernetworks},
  author={Shamsian, Aviv and Navon, Aviv and Fetaya, Ethan and Chechik, Gal},
  booktitle={ICML},
  year={2021}
}

@inproceedings{li2025secure,
  title={Secure On-Device Video OOD Detection Without Backpropagation},
  author={Li, Shawn and Cai, Peilin and Zhou, Yuxiao and Ni, Zhiyu and Liang, Renjie and Qin, You and Nian, Yi and Tu, Zhengzhong and Hu, Xiyang and Zhao, Yue},
  booktitle={ICCV},
  year={2025}
}

@inproceedings{knyazev2023can,
  title={Can we scale transformers to predict parameters of diverse imagenet models?},
  author={Knyazev, Boris and Hwang, Doha and Lacoste-Julien, Simon},
  booktitle={ICML},
  year={2023}
}

@inproceedings{dravid2024interpreting,
  title={Interpreting the weight space of customized diffusion models},
  author={Dravid, Amil and Gandelsman, Yossi and Wang, Kuan-Chieh and Abdal, Rameen and Wetzstein, Gordon and Efros, Alexei and Aberman, Kfir},
  booktitle={NeurIPS},
  year={2024}
}

@inproceedings{
zheng2022symbolic,
title={Symbolic Learning to Optimize: Towards Interpretability and Scalability},
author={Wenqing Zheng and Tianlong Chen and Ting-Kuei Hu and Zhangyang Wang},
booktitle={ICLR},
year={2022}
}

@inproceedings{chen2023symbolic,
  title={Symbolic discovery of optimization algorithms},
  author={Chen, Xiangning and Liang, Chen and Huang, Da and Real, Esteban and Wang, Kaiyuan and Pham, Hieu and Dong, Xuanyi and Luong, Thang and Hsieh, Cho-Jui and Lu, Yifeng and others},
  booktitle={NeurIPS},
  year={2023}
}

@inproceedings{
csordas2020neural,
title={Are Neural Nets Modular? Inspecting Functional Modularity Through Differentiable Weight Masks},
author={R{\'o}bert Csord{\'a}s and Sjoerd van Steenkiste and J{\"u}rgen Schmidhuber},
booktitle={ICLR},
year={2021}
}

@article{bau2020understanding,
  title={Understanding the role of individual units in a deep neural network},
  author={Bau, David and Zhu, Jun-Yan and Strobelt, Hendrik and Lapedriza, Agata and Zhou, Bolei and Torralba, Antonio},
  journal={PNAS},
  year={2020}
}

@inproceedings{qiu2023unlocking,
  title={Unlocking emergent modularity in large language models},
  author={Qiu, Zihan and Huang, Zeyu and Fu, Jie},
  booktitle={NAACL},
  year={2024}
}

@article{zhang2023emergent,
  title={Emergent modularity in pre-trained transformers},
  author={Zhang, Zhengyan and Zeng, Zhiyuan and Lin, Yankai and Xiao, Chaojun and Wang, Xiaozhi and Han, Xu and Liu, Zhiyuan and Xie, Ruobing and Sun, Maosong and Zhou, Jie},
  journal={ACL Findings},
  year={2024}
}

@inproceedings{shevchenko2020landscape, title={Landscape Connectivity and Dropout Stability of SGD Solutions for Over-parameterized Neural Networks}, author={Shevchenko, Alexander and Mondelli, Marco}, booktitle={ICML}, year={2020} }

@inproceedings{tatro2020optimizing, title={Optimizing Mode Connectivity via Neuron Alignment}, author={Tatro, N. Joseph and Chen, Pin-Yu and Das, Payel and Melnyk, Igor and Sattigeri, Prasanna and Lai, Rongjie}, booktitle={NeurIPS}, year={2020} }

@article{balzano2025overview,
  title={An Overview of Low-Rank Structures in the Training and Adaptation of Large Models},
  author={Balzano, Laura and Ding, Tianjiao and Haeffele, Benjamin D and Kwon, Soo Min and Qu, Qing and Wang, Peng and Wang, Zhangyang and Yaras, Can},
  journal={arXiv},
  year={2025}
}

@article{shamsian2024augmentations,
  title={Improved generalization of weight space networks via augmentations},
  author={Shamsian, Aviv and Navon, Aviv and Zhang, David W and Zhang, Yan and Fetaya, Ethan and Chechik, Gal and Maron, Haggai},
  journal={ICML},
  year={2024}
}

@article{ainsworth2022git, title={Git Re-Basin: Merging Models Modulo Permutation Symmetries}, author={Ainsworth, Samuel K. and Hayase, Jonathan and Srinivasa, Siddhartha}, journal={ICLR}, year={2023} }

@inproceedings{yurochkin2019bayesian, title={Bayesian Nonparametric Federated Learning of Neural Networks}, author={Yurochkin, Mikhail and Agarwal, Mayank and Ghosh, Soumik and Greenewald, Kristjan and Hoang, Tin and Khazaeni, Yasaman}, booktitle={ICML}, year={2019} }

@inproceedings{dinh2017sharp,
  title={Sharp minima can generalize for deep nets},
  author={Dinh, Laurent and Pascanu, Razvan and Bengio, Samy and Bengio, Yoshua},
  booktitle={ICML},
  year={2017}
}

@article{gur2018gradient,
  title={Gradient descent happens in a tiny subspace},
  author={Gur-Ari, Guy and Roberts, Daniel A and Dyer, Ethan},
  journal={arXiv},
  year={2018}
}

@inproceedings{kornblith2019similarity, title={Similarity of Neural Network Representations Revisited}, author={Kornblith, Simon and Norouzi, Mohammad and Lee, Honglak and Hinton, Geoffrey}, booktitle={ICML}, year={2019} }

@inproceedings{bansal2021revisiting,
  title={Revisiting model stitching to compare neural representations},
  author={Bansal, Yamini and Nakkiran, Preetum and Barak, Boaz},
  booktitle={NeurIPS},
  year={2021}
}

@inproceedings{li2018measuring, title={Measuring the Intrinsic Dimension of Objective Landscapes}, author={Li, Chunyuan and Farkhoor, Heerad and Liu, Rosanne and Yosinski, Jason}, booktitle={ICLR}, year={2018} }

@inproceedings{ghorbani2019investigation, title={An Investigation into Neural Net Optimization via Hessian Eigenvalue Density}, author={Ghorbani, Behrooz and Krishnan, Shankar and Xiao, Ying}, booktitle={ICML}, year={2019} }

@inproceedings{foret2021sam, title={Sharpness-Aware Minimization for Efficiently Improving Generalization}, author={Foret, Pierre and Kleiner, Ariel and Mobahi, Hossein and Neyshabur, Behnam}, booktitle={ICLR}, year={2021} }

@inproceedings{
ji2020implicit,
title={Gradient descent aligns the layers of deep linear networks},
author={Ziwei Ji and Matus Telgarsky},
booktitle={ICLR},
year={2019}
}

@inproceedings{le2022training, title={Training Invariances and the Low-Rank Phenomenon: Beyond Linear Networks}, author={Le, Thien H. and Jegelka, Stefanie}, booktitle={ICLR}, year={2022} }

@inproceedings{
galanti2024sgd,
title={{SGD} and Weight Decay Secretly Minimize the Rank of Your Neural Network},
author={Tomer Galanti and Zachary S Siegel and Aparna Gupte and Tomaso A Poggio},
booktitle={NeurIPS 2024 Workshop on Mathematics of Modern Machine Learning},
year={2024}
}

@inproceedings{
ongie2019function,
title={A Function Space View of Bounded Norm Infinite Width ReLU Nets: The Multivariate Case},
author={Greg Ongie and Rebecca Willett and Daniel Soudry and Nathan Srebro},
booktitle={ICLR},
year={2020}
}

@inproceedings{
soudry2018implicit,
title={The Implicit Bias of Gradient Descent on Separable Data},
author={Daniel Soudry and Elad Hoffer and Nathan Srebro},
booktitle={ICLR},
year={2018}
}

@inproceedings{
lyu2019gradient,
title={Gradient Descent Maximizes the Margin of Homogeneous Neural Networks},
author={Kaifeng Lyu and Jian Li},
booktitle={ICLR},
year={2020}
}

\end{document}